# Gray Image extraction using Fuzzy Logic

Koushik **Mondal**
Indian Institute of Science Education and Research, Pune, India

Paramartha **Dutta**
Department of Computer & System Sciences
Visva-Bharati University, Santiniketan,
West Bengal, India

Siddhartha **Bhattacharyya**
Department of CS & IT
RCC Institute of Information Technology,
Kolkata, West Bengal, India

**Abstract:**

Fuzzy systems concern fundamental methodology to represent and process uncertainty and imprecision in the linguistic information. The fuzzy systems that use fuzzy rules to represent the domain knowledge of the problem are known as Fuzzy Rule Base Systems (FRBS). On the other hand image segmentation and subsequent extraction from a noise-affected background, with the help of various soft computing methods, are relatively new and quite popular due to various reasons. These methods include various Artificial Neural Network (ANN) models (primarily supervised in nature), Genetic Algorithm (GA) based techniques, intensity histogram based methods etc. providing an extraction solution working in unsupervised mode happens to be even more interesting problem. Literature suggests that effort in this respect appears to be quite rudimentary. In the present article, we propose a fuzzy rule guided novel technique that is functional devoid of any external intervention during execution. Experimental results suggest that this approach is an efficient one in comparison to different other techniques extensively addressed in literature. In order to justify the supremacy of performance of our proposed technique in respect of its competitors, we take recourse to effective metrics like Mean Squared Error (MSE), Mean Absolute Error (MAE), Peak Signal to Noise Ratio (PSNR).

*Keywords: Fuzzy Rule Base, Image Extraction, Fuzzy Inference System (FIS), Membership Functions, Membership values*

## I. Introduction

Over the past few decades, Fuzzy Logic [1] has been used in a wide range of problem domains. Fuzzy Logic is usually regarded as a formal way to describe how human beings perceive everyday concepts. In Fuzzy Image processing fuzzy set theory [2] is applied to the task of image processing. Fuzzy Image Processing is depends upon membership values [3], rule-base and inference engine. The uncertainty in image segmentation and subsequent extraction from noise affected scene effectively handled by Fuzzy Logic. According to [4], fuzzy approaches for image segmentation can be categorized into four classes: segmentation via thresholding, segmentation via clustering, supervised segmentation and rule based segmentation. Among these categories, rule based segmentation are able to take advantage of application dependent heuristic knowledge, and model them in the form of fuzzy rule base. In our case, the heuristic knowledge gathers by the process by already exist threshold segmentation methods helped us to build the rule base. Image Segmentation and subsequent extraction from noise affected scene happen to be crucial phase of image processing. The complex process of human vision is yet not comprehensively explored, in spite of several decades of dedicated study on the problem, may it be from the perspective of basic science or from the viewpoint of research on intelligence. In computer vision, the complex process of recognizing shapes, colors, textures and subsequently grouping them automatically into separate regions or objects within a scene continues to be an open research avenue, intrinsically because of the uncertainty associated with it. Out of the twin objectives of segmentation and extraction, championed earlier, image segmentation appears to be a low-level image processing task that aims at partitioning an image into regions in order that each region/ group consists of homogeneous pixels sharing similar attributes (intensity, colors etc.). The problem becomes even more challenging with the presence of noise in the image scene where elimination of this uncalled for noise component need to be eliminated while preserving the image content as much as possible. Naturally, the extraction of objects prevalent in an image content from a noise affected background. Generally, two steps have

to be considered in order to address any segmentation problem:

Step 1: To formalize the segmentation problem, a mathematical notion of homogeneity or similarity between image regions need to be considered.

Step 2: An efficient algorithm for partitioning or clustering has to be derived particularly to carry the earlier step out in a computationally efficient manner.

The problems of image segmentation become more uncertain and severe when it comes to dealing with noisy images. The vagueness of image information arising out of admixture of the different components has been dealt with soft computing paradigm. Numerous articles and several surveys on gray /monochrome image segmentation techniques have to be reported in this regard [5] [6] [9].

A formal definition of segmentation of an image can be defined as in [7]. Segmentation of image $I$ is a partition $P$ of $I$ into a set of $M$ regions $\{R_m, m=1, 2…M\}$ such that:

1. $\bigcup_{m=1}^{M} R_m = I$ with $R_m \cap R_n = \Phi$,

   $m \neq n, 1 \leq m, n \leq M$

2. $H(R_m) = true \ \forall \ m, 1 \leq m, n \leq M$

3. $H(R_m \cap R_n) = false \ \forall \ R_m$ and $R_n$

   adjacent, $1 \leq m, n \leq M$

Fig. 1: Segmentation definition

Here $H$ is the predicate of homogeneity. A region is homogeneous if all its pixels satisfy the homogeneity predicate defined over one or more pixel attributes such as intensity, texture or color. On the other hand, a region is said to be connected if a connected path exists between any two pixels within the region.

Because of the large diversity of segmentation methods, it is indeed difficult to exhaustively review each individual segmentation techniques up to now. However, segmentation methods can be broadly classified as
(i) region or boundary-based,
(ii) graph-based or
(iii) histogram-based
to mention a few. This paper is presented in the following manner. In the section 2, we would like to discuss survey of recent methodology in this area; section 3 proposes our present work; section 4 clarifies the results and analysis followed by conclusion.

## II. Survey

Gray scale image segmentation approaches are based on either discontinuity and/or homogeneity of gray level values in a region. The approach based on discontinuity, tends to partition an image by detecting isolated points, lines and edges according to abrupt changes in gray levels in two adjacent regions in the scene. The approaches based on homogeneity include thresholding, clustering, region growing and region splitting & merging. Several surveys are reported in the literature to this effect. Fu *et al.* discussed segmentation from the viewpoint of cytology image processing [5]. The paper categorized various existing segmentation techniques into three classes:

i. Characteristic feature thresholding or clustering
ii. Edge detection and
iii. Region extraction.

The segmentation techniques were summarized and comments were provided on the pros and cons of each approach. The threshold selection schemes based on gray level histogram and local properties as well as based on structural, textural and syntactic techniques were described [5][6][7]. Clustering techniques were regarded as "the multidimensional extension of the concept of thresholding". Some clustering schemes utilizing different kinds of features (multi-spectral information, mean/ variation of gray level, texture, color) were discussed. Various edge detection techniques were presented, which were categorized into two classes - parallel and sequential techniques. The parallel edge detection technique[8][9] implies that the decision of whether a set of points is on an edge or not, depends on the gray level of the set and some set of its neighbors, which includes high emphasis on spatial frequency filtering, gradient operators, adaptive local operator, functional approximations, heuristic search and dynamic programming, relaxation and line & curve fitting, while the sequential techniques make decision based on the results of the previously examined points. A brief description of the major component of a sequential edge detection procedure was provided in [5][7]. In those papers region merging, region splitting and combination of region merging and splitting approaches briefly introduced. Haralick *et al.* classified image

segmentation techniques into six major groups [6]:
i. Measurement space guided spatial clustering
ii. Single linkage region growing schemes
iii. Hybrid linkage region growing schemes
iv. Centroid linkage region growing schemes
v. Spatial clustering schemes and
vi. Split & merge schemes.

These techniques are compared on the problem of region merge error, blocky region boundary and memory usage. The hybrid linkage region growing schemes appear to be the best compromise between having smooth boundaries and few unwanted region merges. One of the drawbacks of feature space clustering is that the cluster analysis does not utilize any spatial information. The article also presented some spatial clustering approaches, which combine clustering in feature space with region growing or spatial linkage techniques. It provides a good summary of kinds of linkage region growing schemes. The problem of high correlation and spatial redundancy of multi-band image histograms and the difficulty of clustering using multi-dimensional histograms are also discussed. Sahoo *et al.* surveyed segmentation algorithms based on thresholding and attempted to evaluate the performance of some thresholding techniques using uniformity and shape measures [7]. It categorized global thresholding techniques into two classes:

i. point-dependent techniques (gray level histogram based)
ii. region-dependent techniques (modified histogram or co-occurrence based).

Discussion on probabilistic relaxation and several methods of multi-thresholding techniques was also available in [6][9]. Spirkovska *et al.* regarded image segmentation in a machine vision system as the bridge between a low-level vision subsystem including image processing operations (such as noise elimination, edge extraction etc.) to enhance the image quality on one hand and a high-level vision subsystem including object recognition and scene interpretation on the other [8]. According to [10], Image segmentation approaches are also categorized into four classes:
- pixel based segmentation
- area based segmentation
- edge based segmentation
- physics based segmentation.

Most gray level image segmentation techniques can be extended to color images, such as histogram thresholding, clustering, region growing, edge detection, fuzzy approaches and neural networks. Gray level segmentation methods can be directly applied to each component of a color space. The results can be combined in some way to obtain a final segmentation result. Segmentation may also be viewed as image classification problem based on color and spatial features [9]. Therefore, segmentation methods can be categorized as supervised or unsupervised learning /classification procedures. Power *et. al.* compared different color spaces (RGB, normalized RGB, HSI- hybrid color space) and supervised learning algorithms for segmenting fruit images [12]. Supervised algorithms include Maximum Likelihood, Decision Tree, K-Nearest Neighbor, Neural Networks, etc. Hance *et al.* explored six unsupervised image segmentation approaches [13]:
- Adaptive thresholding
- Fuzzy C-means (FCM)
- SCT/center split
- PCT (Principal Components Transform) median cut
- Split and merge
- Multi-resolution segmentation.

Some algorithms resort to combination of unsupervised and supervised methods to segment color images. Hu *et al.* used unsupervised learning and classification based on the FCM algorithm and nearest prototype rule [14]. The classified pixels are used to generate a set of prototypes, which are optimized using a multilayer neural network. The supervised learning is utilized because the optimized prototypes are subsequently used to classify other image pixels. Eom *et al.* employed a neural network for supervised segmentation and a fuzzy clustering algorithm for unsupervised segmentation [15]. Histogram thresholding is one of the widely used techniques for monochrome image segmentation. It assumes that images are composed of regions distributed with different gray level ranges. The histogram of an image can be separated into a number of peaks (modes) each corresponding to one region and there exists a threshold value corresponding to valley between the two adjacent peaks. However, there is limitation since all the existing thresholding techniques having notional resemblance to gray scale images.

In order to obtain the maximum information between two sources, mode (regions with high densities) and valley (regions with low densities), Guo *et al.* adopted entropy based thresholding method [16]. Mode seeking is decided by the multi-modal probability density function (pdf) estimation and the mode can be found by thresholding the pdf.

A network for classifying an image into distinct regions can be subjected to either supervised or unsupervised learning. The learning would be supervised if external criteria and/or intervention are used and matched by the network output otherwise the learning is unsupervised [13].

Genetic algorithm is another search strategy based on the mechanism of natural selection and group inheritance in the process of biological evolution [17][18]. It simulates the cases of reproduction, mating and mutation in sexual reproduction. GA looks each potential solution as an individual in a group (all possible solutions) and encodes each individual into an encoded domain where the genetic operators [19] can be effectively applied.

### III. Present work

It is interesting that all methods invariably performed poorly for at least one or two instances. Thus it was observed that any single algorithm could not be successful for all noisy image types, even in a single application domain. To obtain the robustness of the thresholding method, we explored the combination of more than one thresholding algorithm based on the conjecture that they could be complementary to each other. The combination of thresholding algorithms can be done at the feature level or at the decision level. At the feature level, we use, for example, some averaging operation on the maximum values obtained from individual algorithms; on the decision level, we have fusion of the foreground-background decisions, for example, by taking the majority decision. Thus it will help us on creating membership envelops in the proposed system.

The algorithm for the proposed work is as follows:

**Step 1.** Read a noisy image as input

**Step 2.** Identify the Region of Interests of the image by different thresholding values

**Step 3.** Extract the image information in terms of pixel attributes and threshold values for future use.

**Step 4.** Construct the different membership envelops of the input image.

**Step 5.** Generate fuzzy rules based on the numerical data obtained from the input image corrupted by noise. The fuzzy rule generation consists of five steps:

a. Discern Input and Output spaces into fuzzy regions
b. Generate fuzzy rules from the given data
c. Map the threshold values obtained from different methods in the corresponding fuzzy region
d. Create a combined fuzzy rule base Determine a mapping on the basis of this combined fuzzy rule base.

**Step 6.** Approximate the value obtained in Step 5

**Step 7.** Display the image constructed thus.

Fuzzy image processing is the collection of all approaches that understand, represent and process the images, their segments and features as fuzzy sets. The representation and processing depend on the selected fuzzy technique and on the problem to be solved. Fuzzy image processing has three main stages: image fuzzification, modification of membership values, and, if necessary, image defuzzification as shown in Fig. 2.

The fuzzification and defuzzification steps are particularly important because of absence of any fuzzy hardware at our disposition. Therefore, the coding of image data (fuzzification) and decoding of the results (defuzzification) are steps indispensible that make possible to process images with fuzzy techniques. The main power of fuzzy image processing lies in the effective use of the middle step (modification of membership values).

### IV. Results and Analysis

The goal of this paper is to describe a linear system using a Mamdani rule base. Specifically, we are modeling the relationship among the images, its extracted counterpart and the fuzzy rule base system using as many as 15 well known thresholding methods. The comparisons are listed in the TABLE I. We evaluated all possible to measure of how well the rule described the actual system behavior over the domain where its antecedent was true. In this paper, we take proper care about how well a Mamdani rule base can be put to model the system, using rules that have high correctness.

The corrupted image, subsequent result obtained by well known methods and proposed fuzzy rule base method are depicted in Fig. 3, Fig. 4 and Fig. 5 respectively.

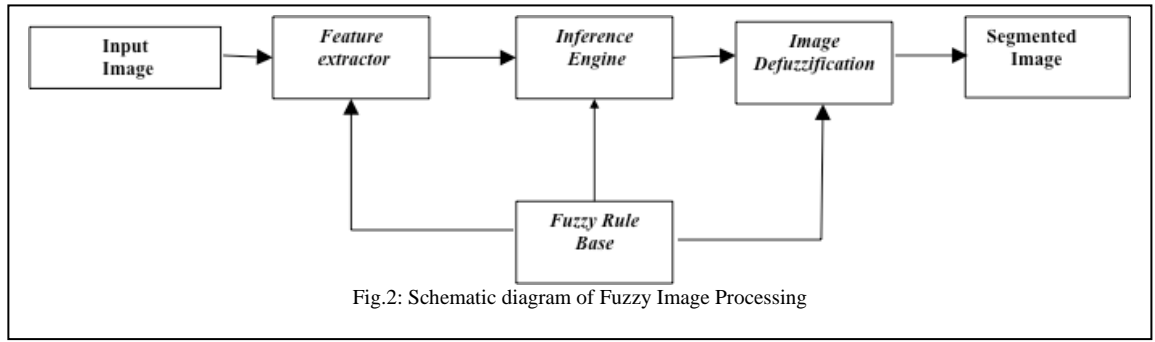
Fig.2: Schematic diagram of Fuzzy Image Processing

TABLE I: Comparison of Gaussian Noise reduction with the help of different thresholding method and our proposed method

| PSNR Calculation for Gaussian Noise | | | | | |
|---|---|---|---|---|---|
| Threshold method/Sigma | 15 | 30 | 45 | 60 | 75 |
| Default | 18.7977 | 18.3826 | 17.8666 | 17.3449 | 16.9336 |
| Huang | 18.6423 | 18.3849 | 17.8578 | 17.3449 | 16.9263 |
| Iso Data | 18.9834 | 18.6435 | 17.4262 | 17.2113 | 16.9321 |
| Li | 18.6785 | 18.3245 | 17.8166 | 17.2354 | 16.8386 |
| Max Entropy | 18.7496 | 18.6748 | 17.4992 | 17.3166 | 16.9392 |
| Mean | 18.6342 | 18.3242 | 17.6822 | 17.3166 | 17.1336 |
| Min Error | 18.9321 | 18.3449 | 17.9321 | 17.8866 | 16.9336 |
| Minimum | 18.2435 | 18.1262 | 17.4221 | 17.3971 | 16.9491 |
| Moments | 18.9213 | 18.6314 | 17.8578 | 17.6314 | 16.937 |
| Otsu | 18.9932 | 18.6808 | 17.5817 | 17.4262 | 16.9336 |
| Percentile | 18.9213 | 18.6314 | 17.4213 | 17.3376 | 16.9321 |
| RenyiEntropy | 18.9491 | 18.6718 | 17.6166 | 17.4262 | 16.9392 |
| Shanbhag | 18.7661 | 18.2381 | 17.9932 | 17.7977 | 17.6166 |
| Triangle | 8.998 | 8.6314 | 7.7262 | 7.3216 | 5.4422 |
| Yen | 18.4262 | 18.2166 | 17.7143 | 17.4132 | 16.9402 |
| **Proposed method** | **29.6435** | **29.3126** | **28.9962** | **28.7143** | **28.6166** |

Fig. 3: Image corrupted by Gaussian Noise

Fig. 4: Image extracted by proposed method

Fig. 5: Image extraction by different thresholding methods

## V. Conclusion

The main features and advantages of this approach are:

i. It provides us a general method to combine measured numerical information into a common framework- a combined fuzzy rule base that theoretically entertains both numerical and linguistic information

ii. It is a simple and straightforward single pass buildup procedure and hence is devoid of any time consuming iterative training as it happens in a comparable neural network or in a neuro-fuzzy approach

iii. There is a lot of freedom in choosing the membership domains in the said design. In fact, this happens to be one of the fundamental challenges

iv. This can be viewed as very general model free integrated fuzzy system for a wide range of image processing problems where "model free" means no mathematical model is required for the problem; "integrated" means the systems integrates all the reported threshold values that are integrated with the systems for finding ROIs and that can help to design adaptive fuzzy regions; and, "Fuzzy" denotes the fuzziness introduced into the system by linguistic fuzzy rules, fuzziness of data, etc.

There are two criteria used in assessing the quality of images. They are subjective criterion and objective criterion. The subjective criterion relies on human beings' individual judgment and interpretation. Naturally, it is shrouded with the possibility of inconsistency and lacks repeatability, it is also time consuming and expensive. One of the standard ways of subjective measurement is called Mean Opinion Score (MOS), it is very tedious, costly and could not be feasible in real time. It has five scales ranging from 'impairment is not noticeable' (best) to 'impairment is extremely objectionable' (worst). On the other hand, the objective criterion available relies on the result of computing some of the following statistical error based methods dependent on pixels difference. Overall image mean absolute error (MAE), overall image mean square error (MSE), signal-to-noise ratio (SNR), or peak signal-to-noise ratio (PSNR) figure this list. The smaller the MAE (or MSE) or the larger the SNR (or PSNR) is, the higher is the quality of the signal. It is fast and repeatable.

There is no universal theory on image segmentation yet that may be universally applicable in all types of images.. This is because image segmentation is subjective in nature and suffers from uncertainty. All the existing image segmentation approaches are, in the main, ad hoc. They are strongly application specific. In other words, there are no general algorithms vis-à-vis color spaces that are uniformly good for all color images. An image segmentation problem is fundamentally one of psychophysical perception and it is essential to supplement any mathematical solutions by a priori knowledge about the image. The fuzzy set theory has attracted more and more attention in the area of image processing because of its inherent capability of handling uncertainty. Fuzzy set theory provides us with a suitable tool, which can represent the uncertainties arising in image segmentation and can model the relevant cognitive activity of the human beings. Fuzzy operators, properties, mathematics, inference rules have found more and more applications in image segmentation. Despite the computational cost, fuzzy approaches perform comparable to or better than their crisp counterparts. The more important advantage of a fuzzy methodology lies in that the fuzzy membership function provides a natural means to model the uncertainty prevalent in an image scene. Subsequently, fuzzy segmentation results can be utilized in feature extraction and object recognition phases of image processing and subsequent computer vision. Fuzzy approach also provides a promising means for color image segmentation.